\documentclass[letterpaper, 10 pt, journal, twoside]{IEEEtran} 
\IEEEoverridecommandlockouts    % Needed for \thanks command
\newcommand\xu[1]{\textcolor{black}{#1}}
\newcommand\xutocheck[1]{\textcolor{black}{#1}}
\newcommand\gui[1]{\textcolor{black}{#1}}
\newcommand\guitocheck[1]{\textcolor{black}{#1}}

\usepackage{glossaries}

\newacronym{mav}{UAV}{Unmanned Aerial Vehicle}
\newacronym{uav}{UAV}{Unmanned Aerial Vehicle}
\newacronym{ovc}{OVC}{Open Vision Computer}
\newacronym{lidar}{LiDAR}{Light Detection and Ranging}
\newacronym{vio}{VIO}{visual-inertial odometry}
\newacronym{satnav}{GNSS}{Global Navigation Satellite Systems}
\newacronym{dgps}{DGPS}{Differential GPS}
\newacronym{rtk}{RTK}{Real-Time Kinematics}
\newacronym{ppk}{PPK}{Post-Processed Kinematic}
\newacronym{gpgpu}{GPGPU}{General-Purpose Graphics Processing Unit}
\newacronym{hri}{HRI}{Human-Robot Interaction}
\newacronym{ugv}{UGV}{Unmanned Ground Vehicle}
\newacronym{uwb}{UWB}{Ultra Wideband}
\newacronym{svm}{SVM}{Support Vector Machine}
\newacronym{fcn}{FCN}{Fully Convolutional Network}
\newacronym{cnn}{CNN}{Convolutional Neural Network}
\newacronym{loam}{LOAM}{LiDAR Odometry and Mapping}
\newacronym{sloam}{SLOAM}{Semantic LiDAR Odometry and Mapping}
\newacronym{slam}{SLAM}{Simultaneous Localization and Mapping}
\newacronym{iot4ag}{IoT4Ag}{NSF Engineering Research Center for the Internet of Things for Precision Agriculture}
\newacronym{grasp-lab}{GRASP Lab}{the General Robotics, Automation, Sensing and Perception Laboratory}
\newacronym{jps}{JPS}{Jump Point Search}
\newacronym{ukf}{UKF}{Unscented Kalman Filter}
\newacronym{sam}{SAM}{Smoothing and Mapping}
\newacronym{icp}{ICP}{Iterative Closest Point}
\newacronym{imu}{IMU}{Inertial Measurement Unit}
\newacronym{tsdf}{TSDF}{Truncated Signed Distance Field}
\newacronym{esdf}{ESDF}{Euclidean Signed Distance Field}
\newacronym{rrt}{RRT}{A rapidly exploring random tree}
\newacronym{fpv}{FPV}{First-person View}
\newacronym{dnn}{DNN}{Deep Neural Network}

\usepackage[table,usenames,dvipsnames]{xcolor}      % color
\usepackage{extarrows}                              % http://ctan.org/pkg/extarrows
\usepackage{enumitem}
\usepackage{wrapfig}

\usepackage{amsmath,amssymb,amsfonts,amsthm,dsfont} % math
\usepackage{algorithm,algorithmicx,listings}        % algorithms
\usepackage[noend]{algpseudocode}			              % necessary for algorithmicx
\makeatletter
\def\BState{\State\hskip-\ALG@thistlm}
\makeatother

\usepackage{graphicx}
\usepackage{tabularx}
\usepackage{subcaption}
\usepackage{textcomp}
\usepackage[font={small}]{caption}   %onehalfspacing
\usepackage[font={small}]{subcaption}
\usepackage[breaklinks=true, colorlinks, bookmarks=true, citecolor=Black, urlcolor=Violet,linkcolor=Black]{hyperref}

\def\argmin{\mathop{\arg\min}\limits}	%
\usepackage[normalem]{ulem}
\usepackage[utf8]{inputenc}
\usepackage[english]{babel}
\usepackage{hyperref}
\hypersetup{
    colorlinks=true,
    linkcolor=blue,
    filecolor=magenta,      
    urlcolor=cyan,
}

\def\argmin{\mathop{\arg\min}\limits}	%
\newcommand{\mat}[1]{\boldsymbol{\mathbf{#1}}}

\DeclareMathAlphabet\mathbfcal{OMS}{cmsy}{b}{n}

\newtheorem*{assumption*}{Assumption}

\newtheorem*{problem*}{Problem}

\usepackage{lipsum}
\usepackage[capitalise]{cleveref}

\usepackage{tikz}
\newcommand\copyrighttext{%
  \footnotesize \textcopyright 2022 IEEE. Personal use of this material is permitted.
  Permission from IEEE must be obtained for all other uses, in any current or future
  media, including reprinting/republishing this material for advertising or promotional
  purposes, creating new collective works, for resale or redistribution to servers or
  lists, or reuse of any copyrighted component of this work in other works.}
\newcommand\copyrightnotice{%
\begin{tikzpicture}[remember picture,overlay]
\node[anchor=south,yshift=7pt] at (current page.south) {\fbox{\parbox{\dimexpr\textwidth-\fboxsep-\fboxrule\relax}{\copyrighttext}}};
\end{tikzpicture}%
}

\begin{document}

\title{Large-scale Autonomous Flight with Real-time Semantic SLAM under Dense Forest Canopy}
% Capitalization of the title

% Make room for more info lines in the \author command  
\author{Xu Liu*, Guilherme V. Nardari*, Fernando Cladera Ojeda, Yuezhan Tao, Alex Zhou,  Thomas Donnelly, \\ Chao Qu, Steven W. Chen, Roseli A. F. Romero, Camillo J. Taylor, Vijay Kumar
\thanks{Manuscript received: September 9, 2021; Revised December 21, 2021; Accepted January 30, 2022.}%Use only for final RAL version
\thanks{This paper was recommended for publication by Editor Pauline Pounds upon evaluation of the Associate Editor and Reviewers' comments.)} %Use only for final RAL version
\thanks{*Equal contribution.~ X. Liu, F. Cladera Ojeda, Y. Tao, A. Zhou, T. Donnelly, C. Qu, S. W. Chen, C. J. Taylor, and V. Kumar are with GRASP Laboratory, University of Pennsylvania {\tt\small\{liuxu, fclad, yztao, alexzhou, tomvdon, quchao, chenste, cjtaylor, vijay.kumar\}@seas.upenn.edu}.}
\thanks{G. V. Nardari and R. A. F. Romero are with Robot Learning Laboratory, University
of São Paulo {\tt\small\{guinardari, rafrance\}@usp.br}.}
\thanks{Digital Object Identifier (DOI): 10.1109/LRA.2022.3154047.}
}
% ras.ral.21-2264.e103c79b

% Use only for final RAL version.

% Paper headers 
\markboth{IEEE Robotics and Automation Letters. Preprint Version. Accepted January, 2022} % Use only for final RAL version
{LIU \MakeLowercase{\textit{et al.}}: Large-scale Autonomous Flight with Real-time Semantic SLAM under Dense Forest Canopy}  
\maketitle
\copyrightnotice

%%%%%%%%%%%%%%%%%%%%%%%%%%%%%%%%%%%%%%%%%%%%%%%%%%%%%%%%%%%%%%%%%%%%%%%%%%%%%%%%
\begin{abstract}
Semantic maps represent the environment using a set of semantically meaningful objects. This representation is storage-efficient, less ambiguous, and more informative, thus facilitating large-scale autonomy and the acquisition of actionable information in highly unstructured, GPS-denied environments. In this letter, we propose an integrated system that can perform large-scale autonomous flights and real-time semantic mapping in challenging under-canopy environments. We detect and model tree trunks and ground planes from LiDAR data, which are associated across scans and used to constrain robot poses as well as tree trunk models. The autonomous navigation module utilizes a multi-level planning and mapping framework and computes dynamically feasible trajectories that lead the UAV to build a semantic map of the user-defined region of interest in a computationally and storage efficient manner. A drift-compensation mechanism is designed to minimize the odometry drift using semantic SLAM outputs in real time, while maintaining planner optimality and controller stability. This leads the UAV to execute its mission accurately and safely at scale.

\begin{IEEEkeywords}
Aerial Systems: Perception and Autonomy, Field Robotics, Robotics and Automation in Agriculture and Forestry, SLAM
\end{IEEEkeywords}

\end{abstract}

    \section{Introduction}
\label{sec:intro}

\IEEEPARstart{A}{utonomously} \xu{surveying large-scale environments and building semantic maps are key capabilities for autonomous mobile robots in many applications such as precision agriculture, infrastructure inspection, and search and rescue.} Semantic maps of forests encode actionable information such as timber volume, yield estimation and forecasting. \glspl{uav} have unique advantages for such tasks: they can hover and fly fast in 3D environments, and they are not as affected by undergrowth or terrain elevation changes as ground robots.

Prior research and commercial solutions using \glspl{uav} for precision agriculture and forestry mostly focus on overhead flight through wide-open spaces. This simplifies operations, but limits what is possible to measure. On the other hand, it is impractical for human pilots to gather data on a large scale under the forest canopy, especially considering the communication range limit in radio and first-person view systems.
Thus, \glspl{uav} capable of autonomous under-canopy flights are indispensable for acquiring such data at scale.

\begin{figure}[t!]
        \centering
            \includegraphics[trim=0 110 0 80, clip, width=0.40\textwidth]{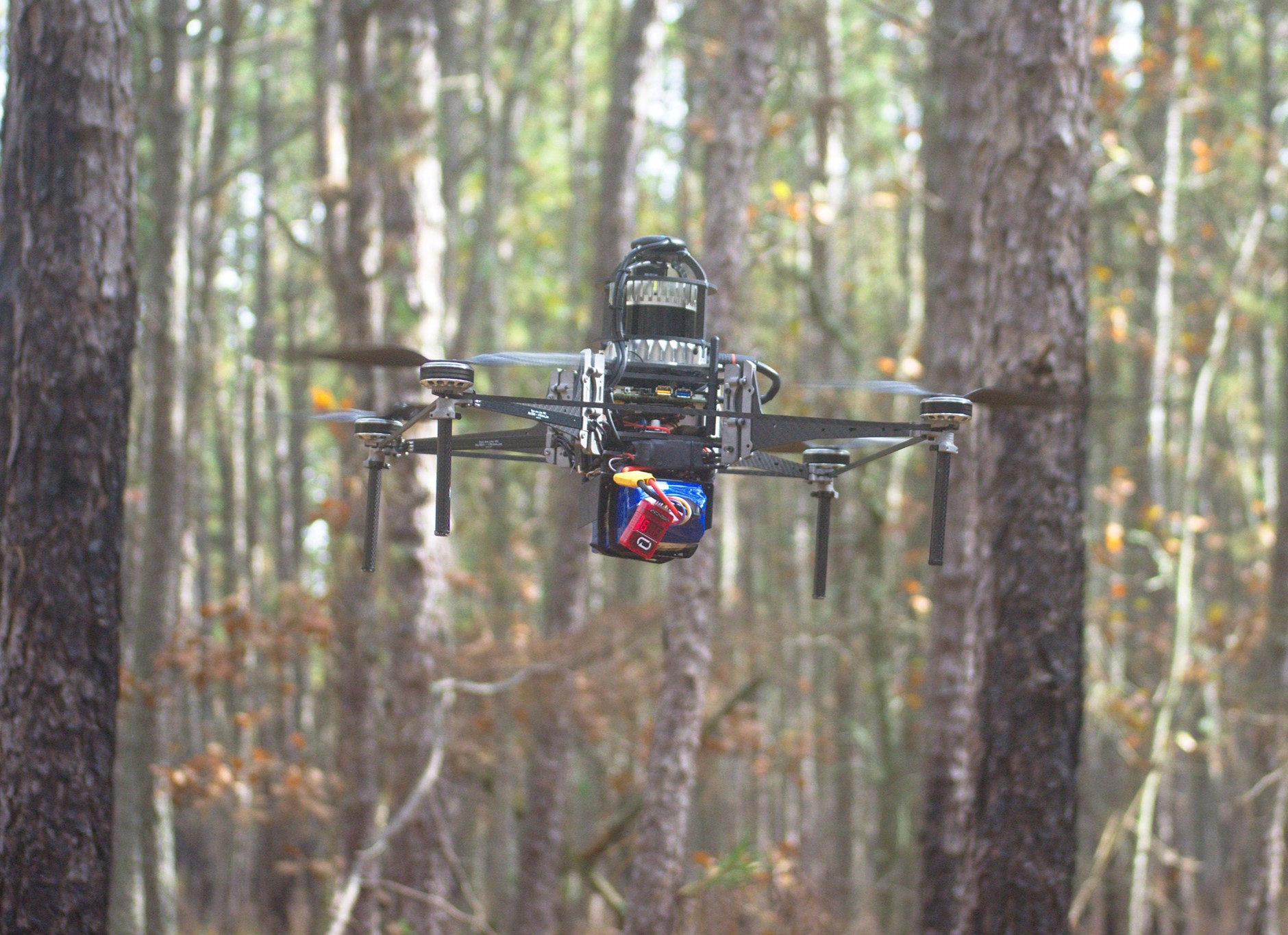}
            \caption{\textbf{Falcon 4 UAV platform}. Our Falcon 4 platform used for this work is the successor of our Falcon 450 platform~\cite{mohta2018experiments} and is equipped with a 3D LiDAR, an \gls{ovc} 3~\cite{quigley2019open} which has a hardware synchronized IMU and stereo cameras, an Intel NUC onboard computer, and a Pixhawk 4 flight controller. The platform has a total weight of 4.2 kg and a 30-minute flight time.}
    \label{fig:uav-platform}
    \vspace{-0.25in}
\end{figure}

\begin{figure*}[th]
    \vspace{-0.25in}
        \centering
            \includegraphics[trim=0 25 0 15, clip, width=0.99\textwidth]{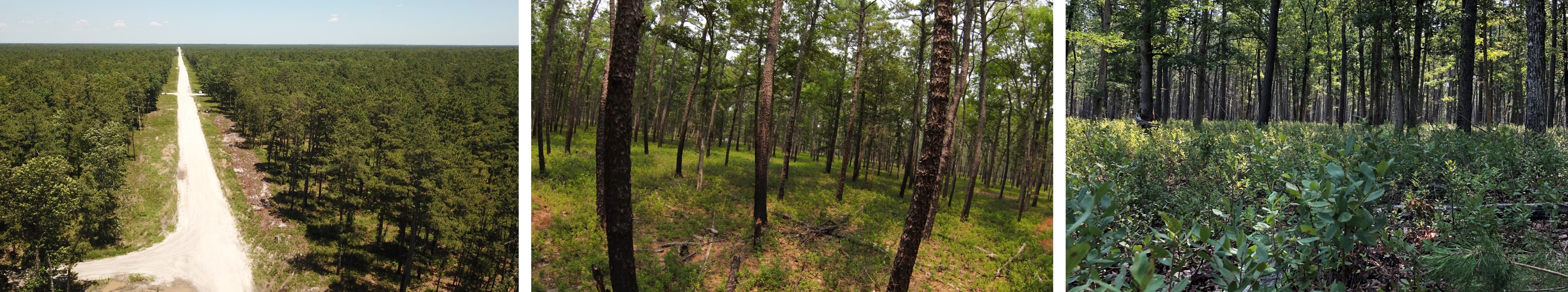}
            \caption{\textbf{Area where the experiments were performed (Wharton State Forest)}. \uline{The left panel} is a view from an over-canopy \gls{uav}, which shows the scale of the environment and the thick tree canopy; %that causes GPS to be unreliable;
            \xu{\uline{The middle panel} is a view from an under-canopy \gls{uav}, which illustrates that the environment is highly unstructured and cluttered; \uline{The right panel} shows that the ground is covered by undergrowth which moves with the wind and is unevenly illuminated. These factors cause traditional VIO or LIDAR odometry systems to drift over time.}}% and drive the \gls{uav} unstable.}
            \vspace{-0.25in}
    \label{fig:show-forest-env}
\end{figure*}

Therefore, developing an autonomous \gls{uav} system that can operate under the tree canopy while extracting actionable information on a large scale is critical for precision agriculture and forestry. This task, however, is still challenging since:
% \begin{enumerate}
    (1) \xu {GPS has unsatisfactory accuracy under dense forest canopies~\cite{forest-gps3-carreiras2013estimating}. Differential or RTK GPS requires reliable communication with a base station, which is impractical for long-range operation in under-canopy environments;}
    (2) the environment is unstructured and dynamic (e.g. leaves blowing in the wind), which is challenging for traditional \gls{slam} algorithms that rely on static geometric features; 
    (3) the environment is highly cluttered, which requires \xu{a capable and reliable navigation system}. 
% \end{enumerate}

\xutocheck{We integrate autonomous flight with semantic \gls{slam} to address these challenges}. %that can perform long-range missions and real-time semantic mapping under a dense forest canopy. 
The main \textbf{contributions} of our paper are: (1) \gui{To our knowledge, this is the first system that integrates semantic SLAM in real-time into an autonomous \gls{uav} feedback loop, while relying only on onboard sensing and computation. The semantic SLAM, coupled with a drift-compensation mechanism minimizes the \gls{uav} odometry drift.} (2) \xutocheck{We propose a \gls{uav} hardware and software system} that is capable of long-range autonomous flights in large-scale, unstructured, cluttered, and GPS-denied environments. {The customized UAV hardware platform (\cref{fig:uav-platform}) has over 30 minutes flight time. Our autonomous flight stack and semantic SLAM algorithm are released at, respectively:

\noindent \url{https://github.com/KumarRobotics/kr_autonomous_flight}

\noindent \url{https://github.com/KumarRobotics/sloam}}

The rest of the paper is arranged as follows: Sec. \ref{sec:related work} surveys related work, followed by a problem statement in Sec. \ref{sec:problem_formulation}. Sec. \ref{sec:proposed_approach_uav_platform} presents our UAV platform. Sec. \ref{sec:proposed_approach_state_estimation} and Sec. \ref{sec:proposed_approach_perception} present the state estimation and perception modules. Sec. \ref{sec:proposed_approach_planning_and_control} presents the planning and control modules, and explains how the integration between autonomy and semantic SLAM is achieved using the drift-compensation module.
Sec. \ref{sec:result_and_analysis} shows our results and analysis in both simulated (\ref{sec:result_and_analysis_simulation}) and real-world experiments (\ref{sec:result_and_analysis_real_world}). Sec. \ref{sec:conclusion} presents the conclusion. A demo video can be found at: \url{https://youtu.be/Ad3ANMX8gd4}.

    \section{Related Work}
\label{sec:related work}

\xu{In this section we provide an overview of related literature. Compared with geometric features, semantic features are usually sparser, less ambiguous and repetitive, and thus can improve robot localization~\cite{bowman2017probabilistic}. In addition, semantic maps are more informative and descriptive, facilitating high-level task or global planning \cite{kostavelis2015semantic}. Generic shape models, such as dual quadrics ~\cite{nicholson2018quadricslam} and 3D bounding boxes~\cite{yang2019cubeslam} can be used to build semantic maps without being limited to specific semantic classes or requiring prior shape models.}

\xu{Several SLAM algorithms are designed specifically for autonomous navigation. Bavle et al. proposed VPS-SLAM \cite{bavle2020vps}, the horizontal and vertical planes are extracted from semantic objects and maintained in the map. The centroids of the planes are then used to provide additional constraints in the graph SLAM optimization. Even though these approaches demonstrate good performance on pre-collected datasets, they are either without autonomy \cite{bavle2020vps, murali2017utilizing}, or with autonomy that does not use any feedback from semantic SLAM \cite{ok2019robust}.} 

\xutocheck{Bavle et al. \cite{bavle2018stereo} propose a particle-filter-based approach that combines visual, inertial and semantic information for \gls{uav} localization, which is then used for navigation. However, their system is demonstrated in small-scale indoor environments with an average flight speed of 0.3 m/s. The semantic objects are not modeled but treated as 3D point landmarks. Most importantly, the UAV relies on a ground station for computation, which means it does not achieve independent autonomy, and is not suitable for outdoor large-scale applications.}

\xutocheck{Several autonomous flight systems that can handle relatively complicated environments with onboard sensing and computation have been proposed. 
Lin et al. described an autonomous UAV system that relies only on a monocular camera and an \gls{imu} to estimate UAV poses and build a dense \gls{tsdf} map in real time \cite{lin2018autonomous}. 
Oleynikova et al. present an autonomous flight system that is capable of exploring and building dense \gls{esdf} maps of unknown environments~\cite{oleynikova2020open}.
Building \gls{esdf} maps is computationally demanding but usually required for planning algorithms that use gradient-based optimization. Zhou et al. propose EGO-Planner~\cite{zhou2020ego} that bypasses this requirement, although an occupancy grid map is still needed.
However, these systems are demonstrated in environments that are less cluttered and have smaller scale than our test environments (\cref{fig:show-forest-env}).} 
%The physical constraints of these platforms, such as battery life, flight speed, and perception range, make them unsuitable for large-scale mapping of complex environments. 
More importantly, none of them incorporate semantic information. The geometric-based odometry algorithms accumulate large drift in under-canopy environment, and the storage demand for maintaining a dense 3D volumetric map is high.

Therefore, developing an autonomous UAV navigation system that requires no external infrastructure and can integrate semantic SLAM into its feedback loop in real time is of key importance for large-scale surveying and mapping missions.

We build upon the system proposed by Mohta et al.~\cite{mohta2018experiments, mohta2018fast}, and make the following improvements: (1) We integrate semantic SLAM into the autonomy stack in real time, which produces more reliable, drift-minimized robot localization, as well as a semantically meaningful map. A drift compensation mechanism is designed to achieve this. (2) We introduce a multi-level planning and mapping framework to improve computation and storage efficiency and support large-scale coverage and mapping missions.

In addition, \gui{we modify the semantic lidar odometry and mapping framework (SLOAM) proposed in Chen et al.~\cite{chen2020sloam} by: (1) employing RangeNet++~\cite{milioto2019iros-rangenet++}, a neural network designed for \gls{lidar} data, giving more accurate segmentation results; (2) integrating the state estimation step with a \gls{vio} algorithm to increase robustness in case of sensor or segmentation failure; and (3) splitting the least-squares pose optimization into two steps to improve robustness.}

\begin{figure*}[t!]
\centering
    % \vspace{-0.10in}
\includegraphics[trim=0 10 0 0, clip, width=6.25in]{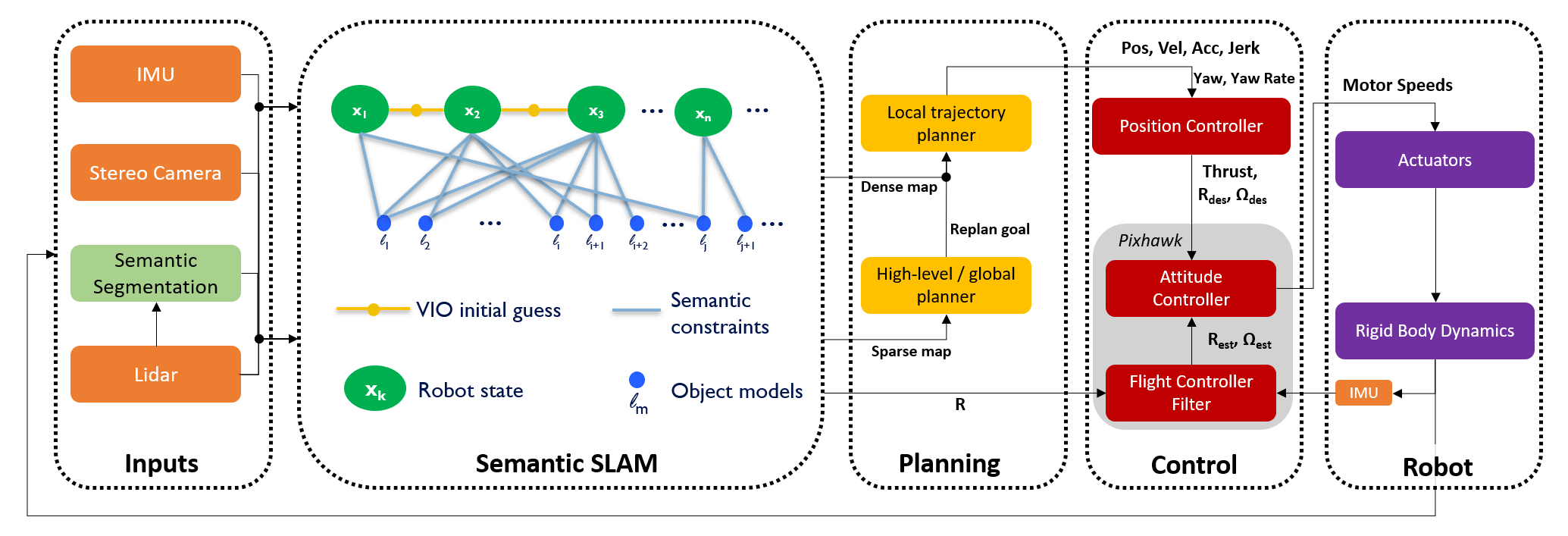}
    \caption{\textbf{Software architecture}. The system uses information from multiple sensors including 3D LIDAR, stereo cameras, and an IMU. It first detects objects in LiDAR scans using RangeNet++~\cite{milioto2019iros-rangenet++}. Object detections are then tracked across the data sequence. This semantic data association is converted into constraints on both the robot poses as well as the object landmark models using the semantic SLAM module, where the relative motion estimated by a stereo VIO algorithm~\cite{sun2018robust} is used as an initial guess, and SLOAM~\cite{chen2020sloam} is used to further optimize the UAV pose and landmarks. The output of the semantic SLAM module is used to reduce the odometry drift through a drift-compensation mechanism. This information is then used by the planning module in real-time. \xu{The high level planner uses a boustrophedon-based coverage planning algorithm~\cite{bahnemann2021revisiting} to cover a polygon region. The mid-level global planner relies on a sparse map to plan the shortest collision-free path to each coverage waypoint}, and the local trajectory planner uses a small but finely discretized robot-centric map to plan a dynamically feasible and collision-free trajectory. Meanwhile, a UKF that runs at 200 Hz is used for the low-level control loop.} 
    \label{fig:realtime-sloam}
    \vspace{-0.20in}
\end{figure*}

\section{Problem Statement}

\label{sec:problem_formulation}

Let $\mathcal{G}$ be a mission defined by the ordered set $\mathcal{G} \triangleq \{g_0, g_1, \dots, g_n\}$, where each $g_i \in \mathcal{G}$ is a waypoint relative to the robot's takeoff position.
The user can either directly specify $\mathcal{G}$ or draw a region of interest using a set of polygons, and the coverage planner will automatically compute the optimal $\mathcal{G}$ to cover this region. Without a prior map, the \gls{uav} must autonomously navigate through the environment and visit all waypoints while detecting and modeling the semantic objects observed. For our current application to forestry, the semantic objects are tree trunks and ground planes.

    \section{Proposed Approach}
\label{sec:proposed_approach}

In this section, we will introduce the individual modules in our system. The system diagram is shown in \cref{fig:realtime-sloam}.

\subsection{UAV platform}
\label{sec:proposed_approach_uav_platform}
The Falcon 4 UAV platform used for our experiments is custom built with a carbon fiber frame. It is equipped with various onboard computers and sensors, as shown in \cref{fig:uav-platform}. 

\xu{Our sensor choices are based on thorough analysis and real-world verification. Cameras and IMUs are inexpensive, lightweight, and have high update rates. However, in under-canopy environments as shown in \cref{fig:show-forest-env}, varying illumination in sunlit forests makes it challenging for any~\gls{vio} system. Exposure must be constantly adjusted, which is detrimental for many algorithms that track visual features since this violates the constant brightness assumption. On the other hand, 3D \glspl{lidar}, although expensive, can have dense and accurate range measurements up to several hundred meters while consuming negligible computation compared to stereo or multi-view depth estimation. They are also robust to environments with direct sunlight and varying illumination. These unique advantages make them indispensable for our purposes of large-scale autonomous navigation and accurate semantic mapping in under-canopy environments. To achieve good robustness, mapping accuracy, and odometry update rates, our system combines stereo \gls{vio} and \gls{lidar}.}  The platform has a payload capacity (including battery) of $\sim$3 kg and can be reconfigured to carry additional onboard computers and other mission-specific sensors. Finally, the platform is powered with a 17,000 mAh Li-ion battery and achieves a flight time of $\sim$30 minutes with all onboard sensors and computers running.

    \subsection{State Estimation}
\label{sec:proposed_approach_state_estimation}

\subsubsection{Odometry}
We use the same odometry system proposed in our previous work~\cite{mohta2018experiments}. The innermost loop of our odometry system is an \gls{ukf}. The \gls{ukf} relies on measurements from stereo \gls{vio} \cite{sun2018robust} and runs at 200 Hz, outputting smooth pose estimates. However, as mentioned in \cref{fig:show-forest-env}, there are many challenges for robot localization under the forest canopy. The geometric features used in traditional \gls{vio} or \gls{lidar} odometry systems will thus become unreliable. Therefore, we use semantic SLAM to correct this drift before feeding it to the planning loop. 

\subsubsection{Semantic Lidar Odometry}
\label{subsubsec:lidar-odom}
Assuming that objects make contact with the ground and \gls{lidar} observations will be gravity aligned, an object detected in a \gls{lidar} beam will likely be present in other beams below it. Using this insight, SLOAM~\cite{chen2020sloam} computes a trellis graph from the point cloud with semantic labels, where nodes are organized as slices. Each node represents a group of points of the same tree from the same horizontal \gls{lidar} beam. Weighted edges are calculated based on the Euclidean distance between nodes. Finally, individual trees are given by the shortest paths starting from the first slice of the graph, which is computed using a greedy algorithm.

Using nodes that belong to the same tree, SLOAM computes a cylinder model $\mathbf{c}$ that estimates the diameter and growth direction of the tree trunk. \gui{To represent the ground, we divide the \gls{lidar} reading according to a circular grid defined by the distance and angle of the space around the sensor. Within each grid cell, we retain the lowest points in the $z$ direction. These points define the set of ground features and are used to fit one plane model $\mathbf{\pi}$ per bin.} With these models, the robot can estimate its pose $\mathbf{T}_k$ at each time step $k$ and create a semantic map with important information for timber inventory, such as diameter at breast height and total tree count.

The main problem of relying purely on semantic information is that the state estimation will also fail if object detection fails. In the following sections, we present improvements to the SLOAM framework to increase stability and robustness for large-scale autonomy.

\subsubsection{VIO Integration}

\gls{vio} will drift under the forest canopy because of light variation and the movement of underbrush and tree branches, where most of the features are detected. However, a robust VIO algorithm with high quality sensors should still provide pose estimates that are consistent within short time intervals. On the other hand, SLOAM may fail if not enough trees are detected to constrain the pose estimation.

We use S-MSCKF~\cite{sun2018robust} as an initial guess for pose estimation and run SLOAM to correct the drift over longer intervals. When the VIO odometry estimates a translational movement of at least $0.5$ meters, a new keyframe is created. For a keyframe $k$, SLOAM will detect trees, perform data association, estimate a pose $\mathbf{T}^{\text{SLOAM}}$ and update the semantic map. This integration is especially valuable to the autonomy stack presented in this work. It is essential in saving computational resources, which enables the whole system to run in real-time.

Since SLOAM's data association is based on nearest-neighbor matching, if the motion between two keyframe poses is large, the association may fail or give false positive matches. We can use the relative motion estimated with VIO to initialize SLOAM and perform data association reliably.

Due to drift \xutocheck{in VIO}, the VIO pose $\mathbf{T}^{\text{VIO}}_k$ and SLOAM pose $\mathbf{T}^{\text{SLOAM}}_k$ may be different. To provide SLOAM with an initial guess, at every keyframe $k$, we store a tuple of poses $(\mathbf{T}^{\text{SLOAM}}_k, \mathbf{T}^{\text{VIO}}_k)$. The initial guess of relative motion between keyframes estimated by the VIO is \gui{$\mathbf{T}^{\text{REL}}_{k} = (\mathbf{T}^{\text{VIO}}_{k-1})^{-1} \cdot \mathbf{T}^{\text{VIO}}_{k}$.}
\gui{It can then be combined with the previous SLOAM pose $\mathbf{T}^{\text{SLOAM}}_{k-1}$} \xutocheck{to form $\mathbf{T}^{\text{GUESS}}_{k} = \mathbf{T}^{\text{REL}}_{k} \cdot \mathbf{T}^{\text{SLOAM}}_{k-1}$, which is used to initialize the SLOAM optimization.}

Let the semantic map $\mathcal{M} \triangleq \{\mathcal{L}_{i}\}_{i=1}^{N}$ with landmarks $\mathcal{L}$ be the set of all landmarks detected by SLOAM over the robot trajectory. For a new observation, we extract a submap $\mathcal{S}_k \subseteq \mathcal{M}$
by selecting from the semantic map a set of trees that are close to the current estimated position $\mathbf{T}^{\text{GUESS}}_{k}$ of the robot based on a distance threshold \gui{$\omega$}. That is,

\begin{equation*}
    \mathcal{S}_k = \{ \left \| \mathbf{c}^{m} - \mathbf{T}^{\text{GUESS}}_{k} \right \|^2  < \omega \, : \, \mathbf{c}^{m} \in \mathcal{M} \}.
\end{equation*}

We store references to each tree in a KD-Tree indexed by their 2D position in the map coordinate frame to query the semantic map efficiently.  SLOAM receives the submap $\mathcal{S}_k$, the initial guess $\mathbf{T}^{\text{GUESS}}_{k}$, and a \gls{lidar} reading $\mathcal{P}^{\text{ROB}}_{k}$ in the robot frame. In this formulation, SLOAM will estimate $\mathbf{T}^{\text{SLOAM}}_{k}$ by performing data association between the current \gls{lidar} reading and the submap.%, which is analogous to the mapping step of the original formulation.

%%%%%%%%%%%%%%%%%%%%%%%%%%%%%%%%%%%%%%%%%%%%%%%%%%%

\begin{figure}[t!]

        \centering
            \includegraphics[trim=0 70 0 210, clip, width=0.4\textwidth]{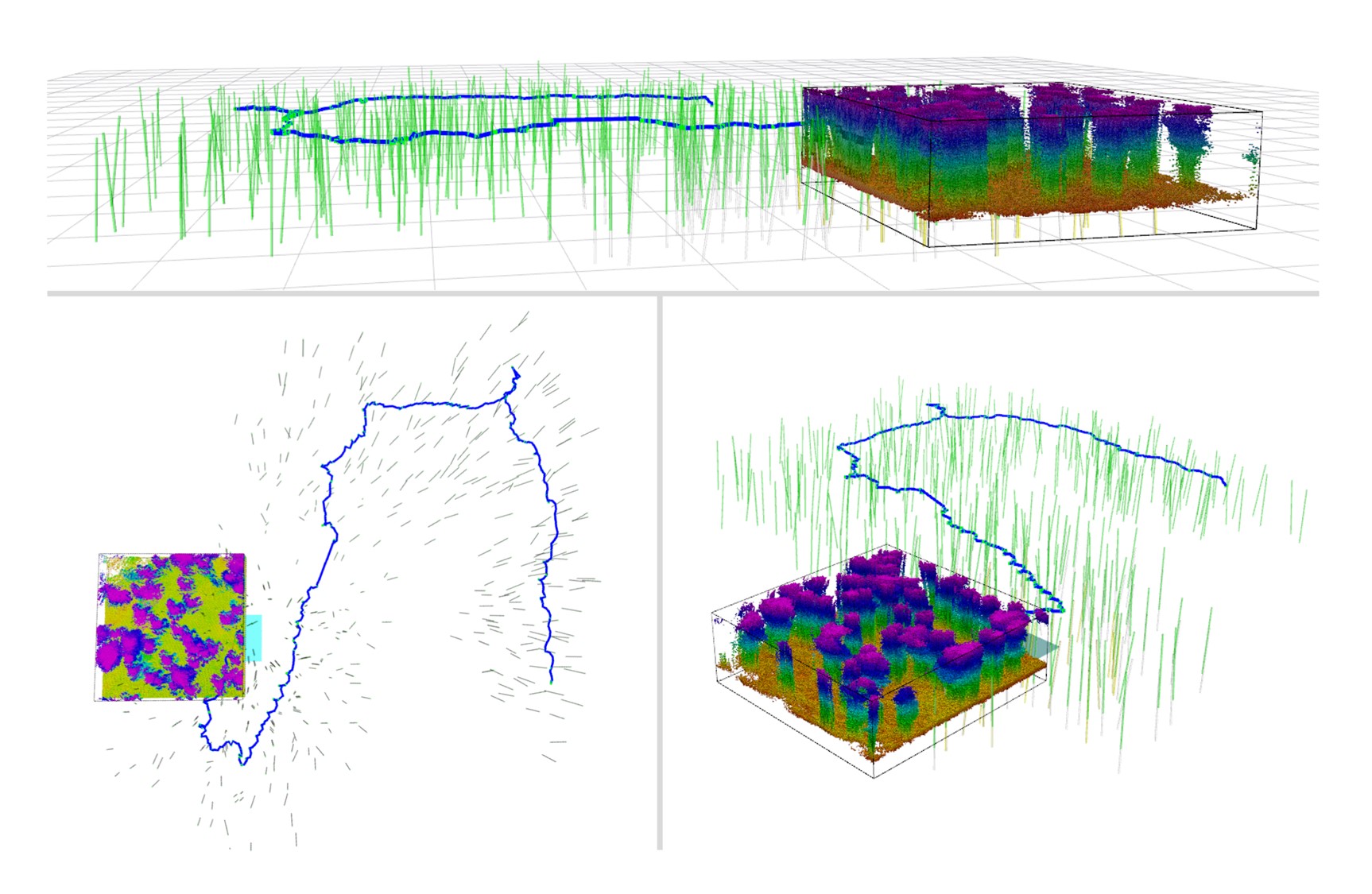}
    \caption{\textbf{Semantic map vs voxel map}. The green cylinders show the semantic map, the black bounding box shows a local voxel map with a 0.1 m resolution, and the blue curve shows the SLOAM-estimated trajectory. It requires 1250 MB to store such a voxel map for a 1 km $\times$ 1 km region. However, storing individual tree models requires $\sim$2 MB for a dense forest of the same size.}
    \label{fig:voxel-vs-semantics}
        \vspace{-0.2in}
\end{figure}

\subsubsection{Lidar State Estimation}

Let $\mathcal{P}^{\text{ROB}}_{k}$ be the \gls{lidar} observation at the current time $t_k$. To perform nearest-neighbor data association between the submap $\mathcal{S}_k$ given an arbitrary robot motion, we transform $\mathcal{P}^{\text{ROB}}_{k}$ using the initial guess $\mathcal{P}^{\text{GUESS}}_{k} = \mathbf{T}^{\text{GUESS}}_{k} \cdot \mathcal{P}^{\text{ROB}}_{k}$. All features and models from this new observation will already be in the initial guess frame.

%%% TREE OPTIMIZATION
Given a set of tree models $\mathcal{L}_{k} \triangleq \{\mathbf{c}_{k}^{i}\}_{i=1}^{N_{k}}$ of size $N_{k}$, for each cylinder $\mathbf{c}_{k}^{i}$, let $\mathcal{T}_{k}^{i} \triangleq \{\mathbf{p}_{j}\}_{j=1}^{\delta_{i,k}}$ be the set of tree feature points of size $\delta_{i,k}$ associated with the landmark. The features in $\mathcal{T}_{k}^{i}$ are associated to the nearest cylinder $\mathbf{c}^{m}$ from the submap $\mathcal{S}_k$. The optimization objective is to minimize the point to cylinder distance of the associated features. Since the robot may not observe the entire tree trunk, the cylinder models are not constrained in the Z direction. For this reason, these models can constrain the X, Y, and Yaw motion. The cost function is given by

\begin{equation}
    \label{eq:pose_optimization_cylinder}
    \begin{aligned}
& \argmin_{\mathbf{T}_{k}^{\text{CYLINDER}}}
& & \sum_{i=1}^{N_{k}}\sum_{j=1}^{\delta_{i,k}}d_{s}(\mathbf{c}_{j}^{m},
 \mat{p}_{j}).\\
\end{aligned}
\end{equation}
where $\mathbf{T}_{k}^{\text{CYLINDER}}$ is an \textbf{SE}(3) transformation with three degrees of freedom (translation Z, Pitch, and Roll are fixed). 

%%% GROUND OPTIMIZATION

\gui{$\mathcal{G}_k \triangleq \{\mathbf{p}_l\}_{l=1}^{\gamma_k}$ defines the set of ground features for the current keyframe. Each feature $\mathbf{p}_l$ is associated to a ground model $\mathbf{\pi}_{l}$ from the previous keyframe based on its euclidean distance to the plane centroid. The optimization objective is to find the transformation that minimizes the point-to-plane distance $d_{\pi}$. Since the ground planes will always be below the robot, it can constrain the Z, pitch, and roll motion. The cost function is given by
}

\begin{equation}
    \label{eq:pose_optimization_ground}
    \begin{aligned}
& \argmin_{\mathbf{T}_{k}^{\text{GROUND}}}
& & \sum_{l=1}^{\gamma_{k}}d_{\pi}(\mat{\pi}_{l}, \mat{p}_{l}),\\
\end{aligned}
\end{equation}
\gui{where $\mathbf{T}_{k}^{\text{GROUND}}$ is an \textbf{SE}(3) transformation with three degrees of freedom (translation X, Y and Yaw are fixed).}
Finally, the SLOAM pose $\mathbf{T}^{\text{SLOAM}}_{k}$ is given by %integration of the initial guess, $\mathbf{T}_{k}^{\text{GROUND}}$ and $\mathbf{T}_{k}^{\text{CYLINDER}}$.

\begin{equation}
    \mathbf{T}^{\text{SLOAM}}_{k} = \mathbf{T}^{\text{GUESS}}_{k} \cdot \mathbf{T}_{k}^{\text{GROUND}} \cdot \mathbf{T}_{k}^{\text{CYLINDER}}.
\end{equation}

% In the original SLOAM formulation, both ground and tree costs were solved jointly and all 6 degrees of freedom were estimated together with non-linear least squares.
\gui{LeGO-LOAM~\cite{shan2018lego} proposes a similar two-step optimization formulation, where the authors report $35\%$ reduction in computation and similar accuracy when compared to estimating the full pose in one problem. LeGO-LOAM uses the output of the ground optimization to constrain the X,Y and Yaw parameter estimation. However, since we have an initial guess from the odometry, we solve \cref{eq:pose_optimization_cylinder} and \cref{eq:pose_optimization_ground} independently. Our primary motivation for splitting the optimization is to increase robustness in failure cases. This formulation can provide constraints to part of the robot pose even in observations where the object detection or data association partially fails.}

\gui{A user-defined threshold controls the minimum number of feature-to-model matches required to perform the optimization. That is, if the number of ground features is low, we can set $\mathbf{T}_{k}^{\text{GROUND}} = I$, where $I$ is the identity transform. If not enough tree features are matched, we can set $\mathbf{T}_{k}^{\text{CYLINDER}} = I$. If both ground and tree features have few matches, SLOAM will use the initial guess to update the map.}

%%%%%%%%%%%%%%%%%%%%%%%%%%%%%%%%%%%%%%%%%%%%%%%%%%%

\subsection{Perception}
\label{sec:proposed_approach_perception}
\subsubsection{Semantic Mapping}
Given a new pose estimate $\mathbf{T}^{\text{SLOAM}}_{k}$, SLOAM will project the models derived from the current sweep, and associate cylinder models to the ones in the submap using a cylinder-to-cylinder distance. These associations update the tree models to the latest estimate, and cylinders that do not have a match are added to the map.

\subsubsection{Semantic segmentation on point clouds}

The original SLOAM~\cite{chen2020sloam} implementation used a lightweight neural network to perform semantic segmentation at \gls{lidar} frequency on the CPU. However, we observed that the segmentation would not be accurate along object edges, considering leaves and small branches as part of the trunk. As explained in section~\ref{subsubsec:lidar-odom}, this is mitigated with the Trellis Graph instance detection. However, it is still challenging to filter these unwanted points that add noise to the cylinder model and, consequently, state estimation. Without the need to run inference on every scan, we improve SLOAM by using RangeNet++~\cite{milioto2019iros-rangenet++}, a more robust architecture designed for \gls{lidar} data.

\subsection{Planning and Control}
\label{sec:proposed_approach_planning_and_control}
\label{sec:hierchical framework and drift correction}

%%%%%%%%%%%%%%%%%%%%%%%%%%%%%%%%%%%%%%%%%%%%%%%%%%%

\begin{figure}[t!]
        \centering
            \includegraphics[trim=0 0 0 0, clip, width=0.425\textwidth]{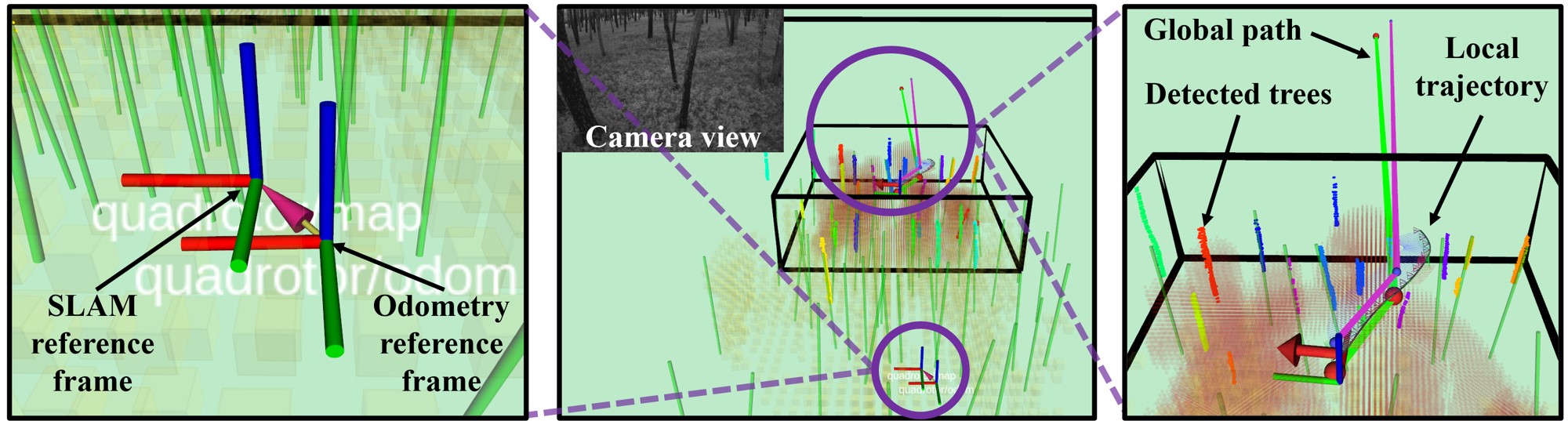}
            \caption{\textbf{Visualization of real-time perception and decision making of the autonomy stack in a real-world forest}. The green cylinders are tree models estimated by SLOAM, the small bounding box is the local voxel map, and the black line shown at the top of the middle panel is the edge of the global voxel map (1 km $\times$ 1 km $\times$ 10 m). \uline{The left panel} shows a zoomed-in view of the SLAM and odometry reference frames. 
            The \gls{vio} and SLOAM estimated poses are used to calculate a transform between the two reference frames, which compensates the \gls{vio} drift.
            %The drift of \gls{vio} is compensated by moving the odometry frame relative to SLAM frame. 
            In \uline{the right panel}, \xutocheck{the drift compensation is used by the planning module to output trajectories that guide the UAV to execute its mission accurately.}}
    \label{fig:rviz-view-autonomous-flight}
    \vspace{-0.2in}
\end{figure}
%%%%%%%%%%%%%%%%%%%%%%%%%%%%%%%%%%%%%%%%%%%%%%%%%%%
Our previous efforts \cite{mohta2018experiments, mohta2018fast} focused on fast autonomous flight in cluttered and GPS-denied environments. However, to autonomously acquire data in dense under-canopy environments that are significantly more cluttered, unstructured, and larger in scale, various challenges need to be addressed. First, we introduce a hierarchical planning and mapping framework to reduce computation and leverage the sparsity of the semantic map to alleviate memory demand. Second, we introduce a real-time drift compensation module to handle the odometry drift correction by semantic SLAM, while maintaining the planner optimality and the controller stability.

\subsubsection{Hierarchical planning and mapping framework}
\label{subsec:hierchical planning and mapping}
We introduce a multi-level mapping and planning framework as illustrated in \cref{fig:rviz-view-autonomous-flight}. Our top-level map consists of semantic models that helps the \gls{uav} to accurately localize itself over long distances and keep track of the information of interest in an efficient manner. Our mid-level (global) map is a large but coarsely discretized voxel map. Our low-level (local) map is a small but finely discretized, robot-centric voxel map, and is updated at the \gls{lidar} rate (10 Hz). The high level planner uses a boustrophedon-based coverage planning algorithm~\cite{bahnemann2021revisiting} to cover the region of interest defined by the user using one polygon to denote the boundary, and additional polygons to denote obstacles or uninterested regions. The user also needs to specify the sensing radius and overlap ratio. The mid-level global planner uses the jump point search algorithm to generate the shortest collision-free and long-horizon path to the next coverage waypoint $g_i$. Each waypoint is parameterized by the quadrotor's flat outputs~\cite{mellinger2011minimum}, i.e., $g_i = [x_i, y_i, z_i, \psi_i]^\mathbf{T}$ where $\mathbf{x}_i = [x_i, y_i, z_i]^\mathbf{T}$ are the 3D positions of the quadrotor's center of mass, and $\psi$ is the yaw angle.% When the distance between the \gls{uav} and $g_i$ is smaller than a given threshold, $g_i$ will be set as reached and $g_{i+1}$ will be used.} 

The local planner uses a motion-primitive-based algorithm, where the 3-rd order derivative of positions (i.e., jerk) in quadrotor's flat outputs are used as control inputs. This algorithm can generate safe and dynamically-feasible trajectories. We refer to the original work \cite{liu2018search} for more details. The local goal is generated by finding the first intersection between the global path and the local map boundaries (the global goal will be used if it lies inside the local map). The planner only needs to reach the local goal within a position tolerance of 4 m, and the velocity is not specified unless the \gls{uav} is close to the final waypoint $g_n$. This avoids unnecessary acceleration and deacceleration, and reduces the replan computation cost.

\subsubsection{Trajectory tracking and control}
To speed up the re-planning process, our trajectory tracker assumes that the \gls{uav} can track the planned trajectory tightly. The assumption is valid since our trajectory is dynamically feasible. By making this assumption, we can reuse the motion primitive graph built during the previous re-plan steps. Since the local planner is running at 2 Hz, the updates in the local map and local goal are usually small. Thus, the planner only needs to expand a small number of additional nodes to reach the new goal, significantly reducing the re-plan time. In addition to tracking the position outputs, if yaw alignment is turned on, the trajectory tracker will use a constant ${\dot{\psi}}$ to align ${{\psi}}$ with the direction of displacement on the x-y plane. For every time step t, the trajectory tracker outputs $\mathbf{s}_t = [{\mathbf{x}}_t^\mathbf{T}, \dot{\mathbf{x}}_t^\mathbf{T}, \ddot{\mathbf{x}}_t^\mathbf{T},\dddot{\mathbf{x}}_t^\mathbf{T}, \psi_t, \dot{\psi}_t]^\mathbf{T}$. $\mathbf{s}_t$ is calculated at 200 Hz. Under extreme circumstances, such as large external disturbance, the \gls{uav} may fail to track the trajectory tightly. Thus, the distance $d$ from the \gls{uav} to the trajectory is constantly monitored. When $d >= 0.5$ m, the state machine will automatically trigger the stopping policy. The stopping policy calculates $\mathbf{s}_t$ using the maximum allowed $\dddot{\mathbf{x}}$ and $\ddot{\psi}$, which brings $\dot{\mathbf{x}}$ and $\dot{\psi}$ to zero as soon as possible while keeping $\ddot{\mathbf{x}}$ within the maximum values. When the \gls{uav} safely stops, the state machine will start a new re-plan process.

A nonlinear geometric-based controller~\cite{lee2010geometric}, which has been demonstrated to have good tracking performance for fast quadrotor flights~\cite{mohta2018fast}, is used as the position controller. Based on $\mathbf{s}_t$ and the odometry, it will calculate the desired thrust, orientation, and body rate at 200 Hz. The Pixhawk 4 flight controller takes in these values and runs an onboard filter and orientation controller to calculate the motor speeds. 

\subsubsection{Real-time drift compensation module}
\label{subsec:real-time drift compensation}
A naive approach to correct the drift induced by the \gls{vio} is to feed the semantic SLAM estimated \gls{uav} poses directly into the planner and controller. However, unlike the filtering-based \gls{vio} that gives smooth pose estimates, the semantic SLAM estimated poses may not be locally smooth (due to, e.g., loop closures). This will cause the \gls{uav} to deviate from the existing trajectory. In such scenario, the motion primitive graph can no longer be reused, thus resulting in a much longer re-plan time.
Even if the re-planning process is triggered when this happens, the new trajectory will not be immediately available due to the re-plan time. Such delay will cause the \gls{uav} to have no feasible trajectory to track, thus resulting in dangerous behaviors.

\guitocheck{A more reliable solution is to maintain separate SLAM and odometry reference frames. Given a SLOAM pose $\mathbf{T}^{\text{SLOAM}}_{k}$, the odometry drift is calculated by $\mathbf{T}^{\text{DRIFT}}_{k} = \mathbf{T}^{\text{SLOAM}}_{k} \cdot (\mathbf{T}^{\text{VIO}}_{k})^{-1}$. The inverse of the drift, which is the transformation from the semantic SLAM estimated pose to the \gls{vio} estimated pose, is applied on the odometry reference frame.}

\guitocheck{The global goal, given by a waypoint $g \in \mathcal{G}$ along the coverage path, lies in the SLAM reference frame and is transformed into the odometry reference frame $g^{\text{ODOM}} = (\mathbf{T}^{\text{DRIFT}}_{k})^{-1} \cdot g^{\text{SLOAM}}$. Using $g^{\text{ODOM}}$ the global planner will output a global path that is drift compensated when re-planning.} \xu{This compensation leads the local planner to generate a drift-compensated local trajectory, while tightly tracking the existing trajectory. The optimality of the local trajectory planner is also maintained.} Thus, the trajectory and control commands will guide the \gls{uav} to execute its long-range mission accurately without frequent stops. \xu{This mechanism also allows our autonomous flight system to directly work with other SLAM or even GPS estimated poses.}

    \section{Results and Analysis}
\label{sec:result_and_analysis}

% \subsection{Overview of experiments}
% what kind of experiment we've done in real world as well as in simulation
\guitocheck{Our system is first developed and tested in simulated environments using the Unity engine\footnote{\url{https://unity.com}}, and then demonstrated and improved in real-world forests. The simulation is especially important for training and testing our \gls{lidar}-based semantic segmentation module as well as the behaviour of the autonomous navigation module without the risk of damaging the~\gls{uav}}. We conduct real-world experiments at the Wharton State Forest, a natural forest located in the U.S. state of New Jersey. It consists of approximately 122,880 acres of pinelands as shown in \cref{fig:show-forest-env}. There is significant variability in tree densities, sizes, shapes, and undergrowth conditions, thus making this forest ideal for verifying our system's robustness and generalizability. 

An important prerequisite is to train the segmentation neural network. We train RangeNet++ on simulator data and then fine-tune on real-world data. It is worth noting that we do not use any labeled data from the Wharton State Forest for training. Instead, we use data collected in a different pine forest in Arkansas, US. We label 50 \gls{lidar} scans to fine-tune the model, 37 for training and 13 for validation. An example of the segmentation result is shown in \cref{fig:rviz-view-autonomous-flight}. We use this model across all experiments without any re-training. It runs at 2 Hz using up to 3 threads of the 12-thread i7-10710U processor of the onboard Intel NUC 10 computer. This is sufficient since semantic segmentation is performed only on keyframes.

\subsection{Simulation experiments and analysis}
\label{sec:result_and_analysis_simulation}

We set up an automatic pipeline that allows us to generate simulated environments based on real-world tree positions that we collected during field trips. We use various high-fidelity tree models that are available in the Unity asset store and customize their sizes, shapes, orientation, as well as undergrowth conditions to simulate the real-world forests.

\begin{figure}[ht!]
        \centering
        % \begin{subfigure}[t]{0.22\textwidth}
        %     \centering
        %     \includegraphics[trim=0 0 0 0, clip, height=0.50in]{imgs/coverage/coverage-executing.png}
        % \end{subfigure}

        \begin{subfigure}[t]{0.25\textwidth}
            \centering
            \includegraphics[trim=0 10 0 10, clip, height=0.50in, angle=-5, origin=c]{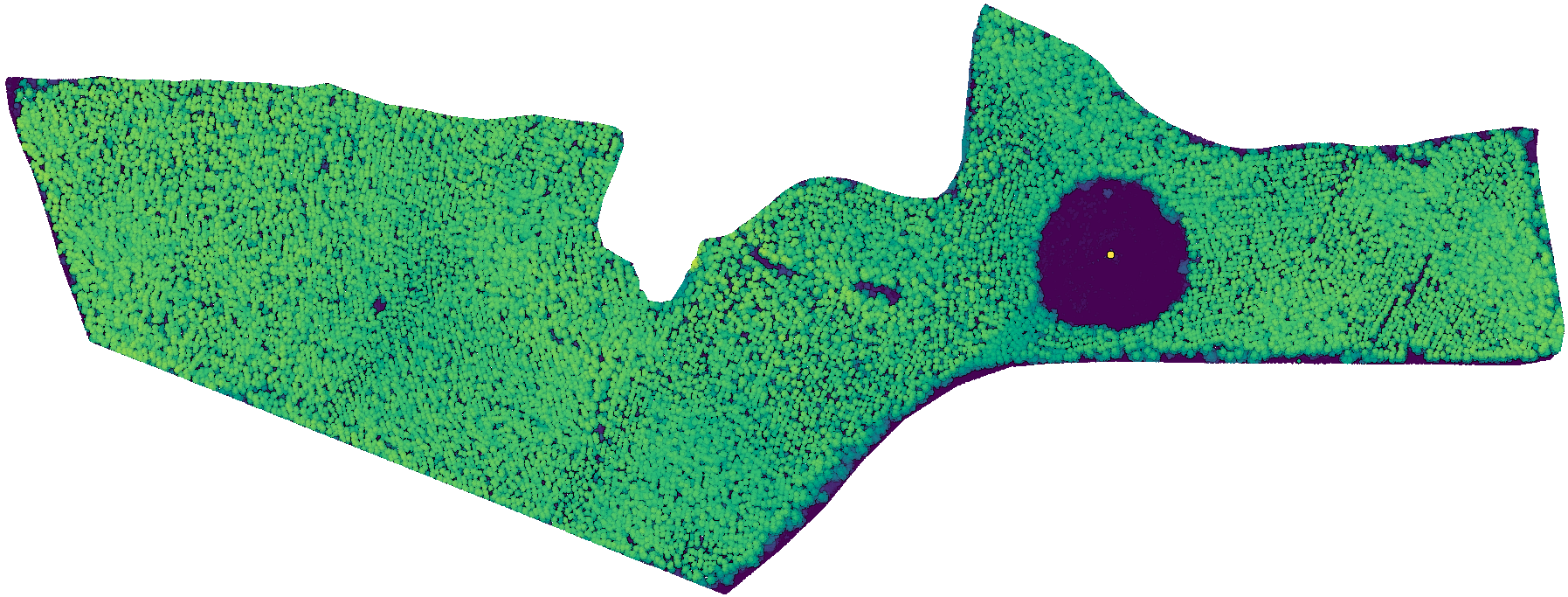}
        \end{subfigure}
        \begin{subfigure}[t]{0.225\textwidth}
            \centering
            \includegraphics[trim=0 10 0 10, clip, height=0.50in, angle=0, origin=c]{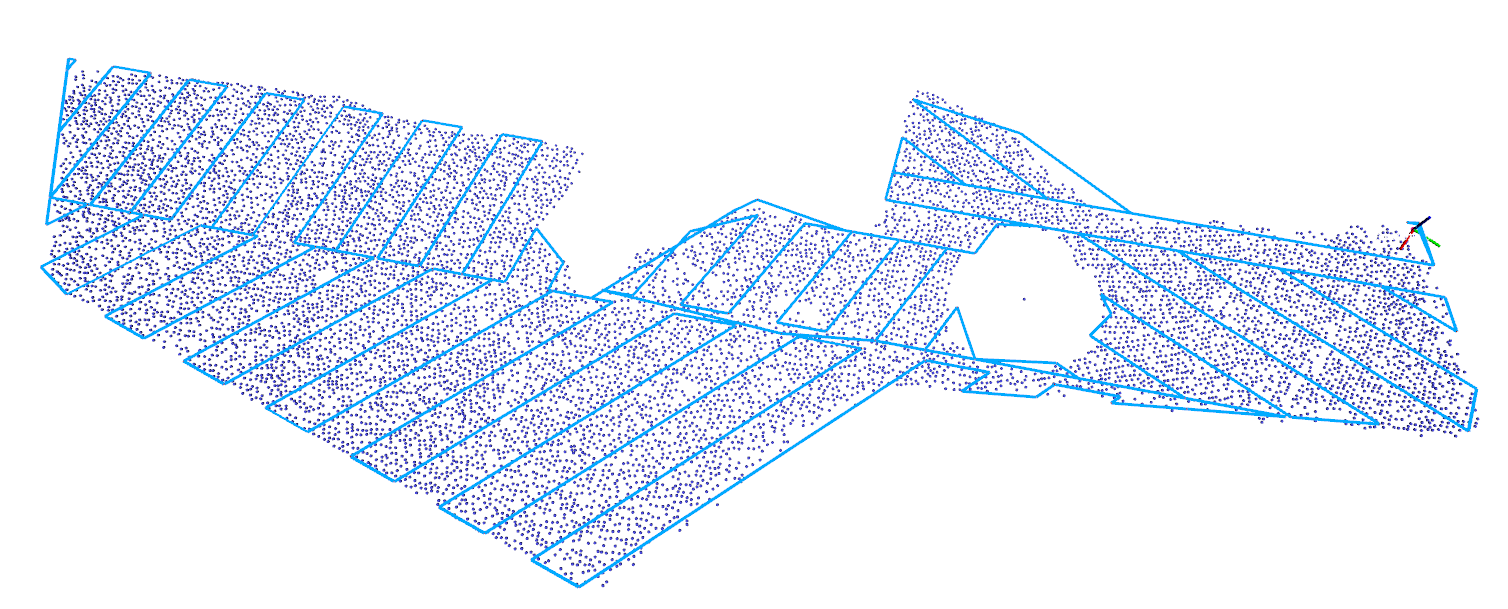}
        \end{subfigure}
        \begin{subfigure}[t]{0.225\textwidth}
            \centering
            \includegraphics[trim=0 0 0 0, clip, height=0.55in, angle=-5, origin=c]{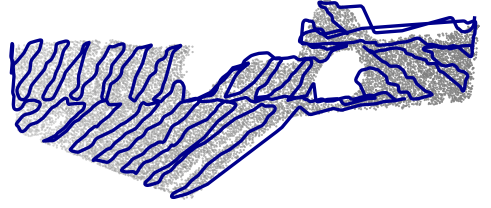}
        \end{subfigure}
        \begin{subfigure}[t]{0.225\textwidth}
            \centering
            \includegraphics[trim=0 0 50 0, clip, height=0.5in, angle=0, origin=c]{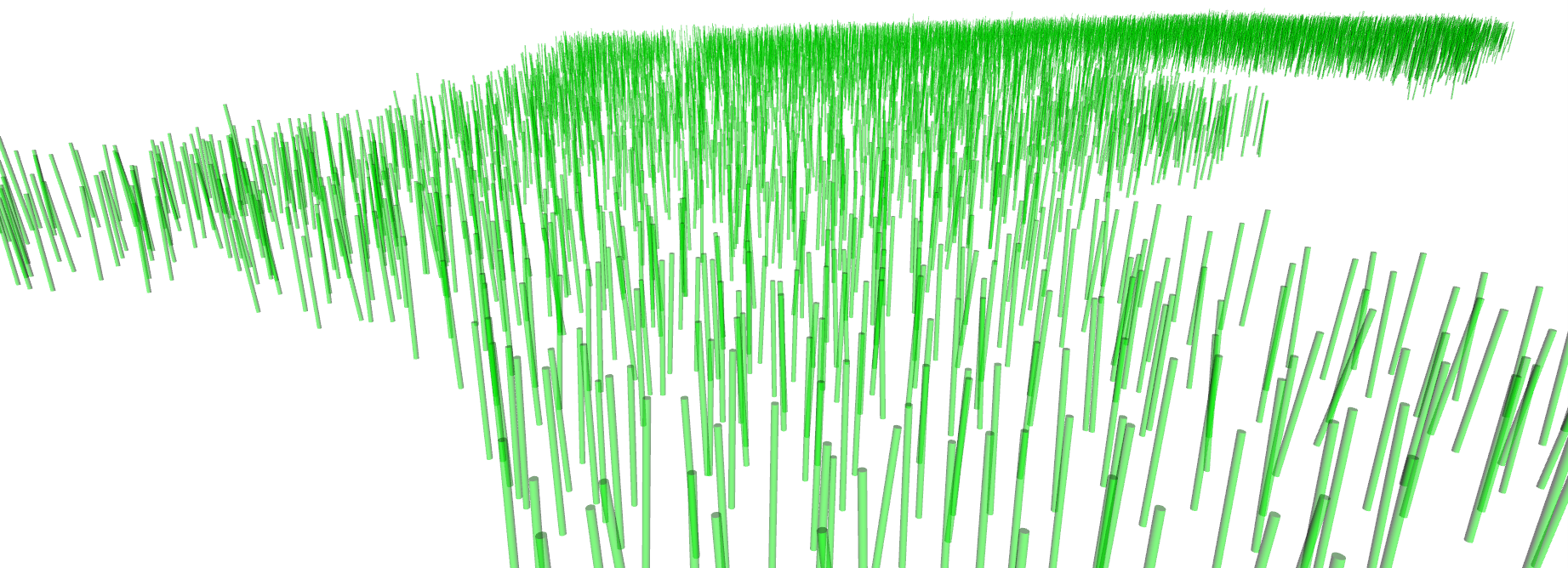}
        \end{subfigure}

    \caption{\xutocheck{\textbf{Large-scale coverage experiment}. The upper left panel shows the overhead \gls{lidar} data of a 0.77 km$^2$ forest stand that is used to create the simulated environment. The upper right panel shows the automatically generated coverage plan based on a user-defined coverage region. The UAV is able to finish this mission autonomously within 1 hour, without any human intervention. The trajectory length is $\sim$10 km and the average flight speed is $\sim$2.9 m/s. The lower left panel shows the semantic map generated by SLOAM consisting of 10,897 tree models, where the semantic segmentation uses the same model as we use for the physical experiments. The lower right panel shows a close-up of the semantic map.}}
    \label{fig:coverage-planning}
        \vspace{-0.20in}
\end{figure}

\begin{figure}[!ht]
        \vspace{-0.05in}
        \centering
        
        \begin{subfigure}[t]{0.24\textwidth}
            \centering
            \includegraphics[trim=0 0 0 0, clip, height=1.2in]{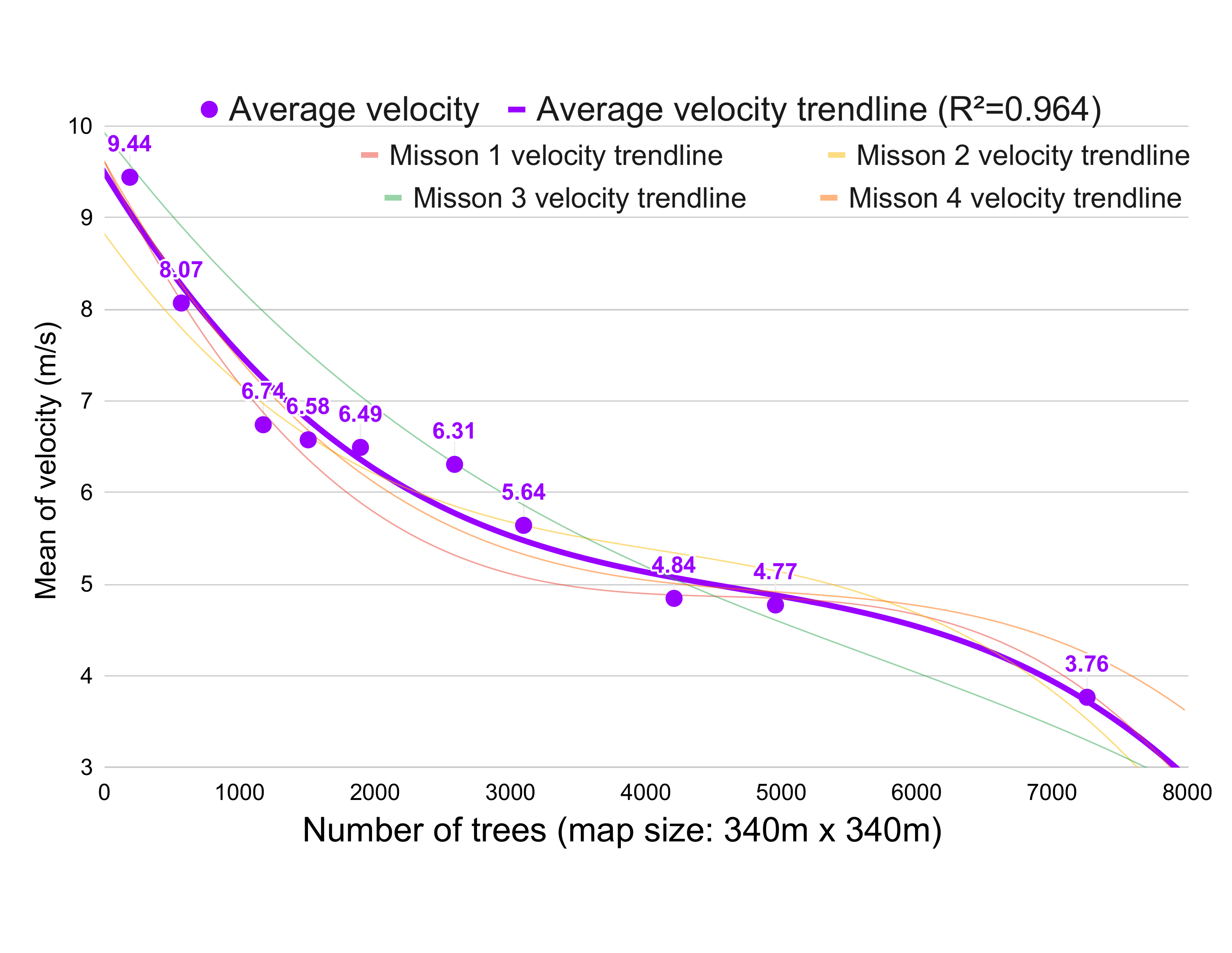}
            \caption{Average flight velocity}
            \label{fig:density-vs-velocity1}
        \end{subfigure}
        \begin{subfigure}[t]{0.24\textwidth}
            \centering
            \includegraphics[trim=0 0 0 0, clip, height=1.2in]{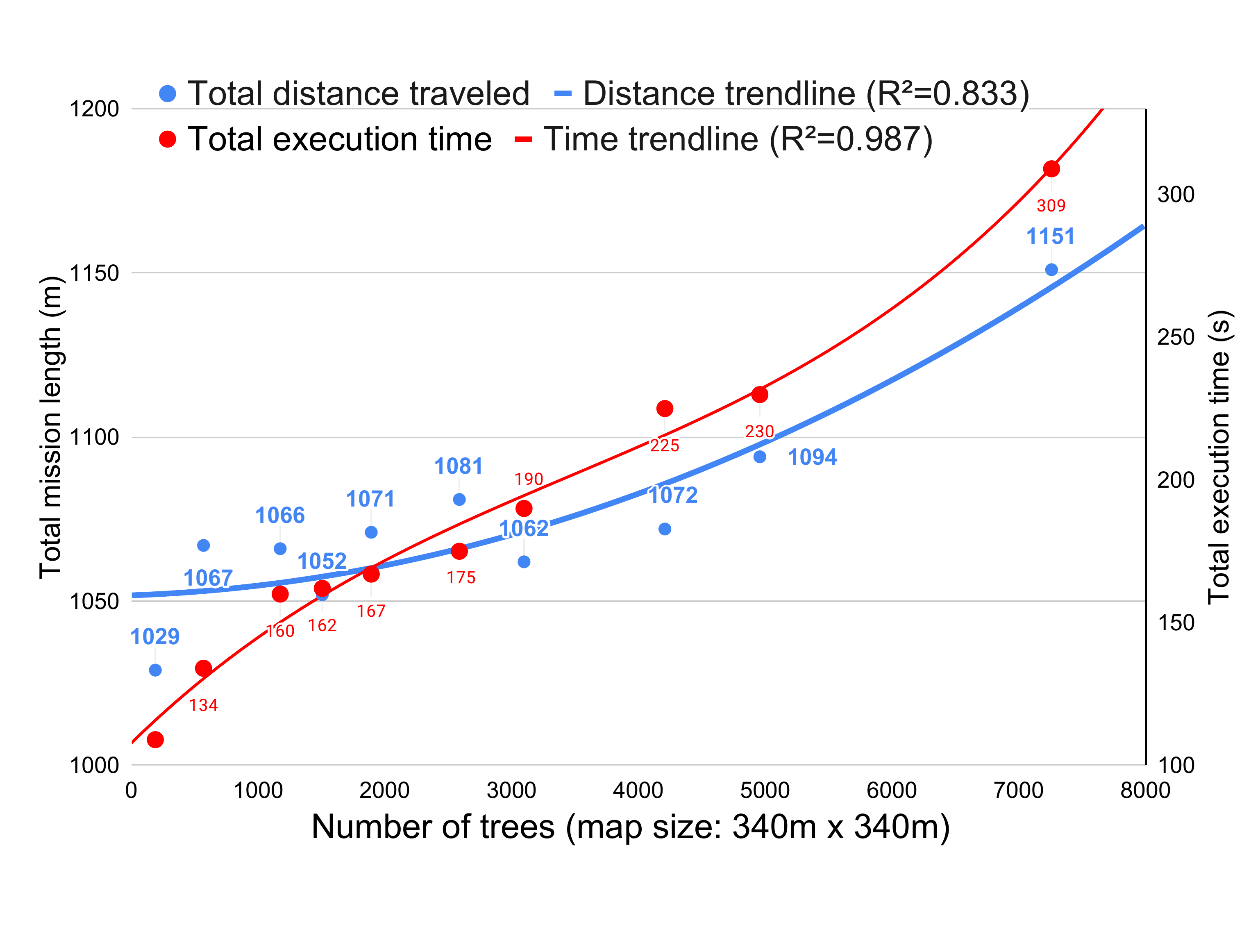}
            \caption{Travel distance and time}
            \label{fig:density-vs-length}
        \end{subfigure}
        \caption{\textbf{Autonomous flight system performance v.s. tree density.} 10 environments with varying densities are used for this set of experiments. For each environment, 4 flight missions are executed. These missions' total straight-line length is $\sim$1 km. On the left panel, the purple curve shows that average flight velocity decreases as the tree density increases. On the right panel, the blue curve shows that the distance traveled by the robot increases as the tree density increases. These two combined lead to a more drastic increase in time to complete the mission, as shown by the red curve on the right.}
        \vspace{-0.2in}
\end{figure}

% large scale + semantic mapping
\gui{In the first simulated experiment, we perform large-scale coverage. Using \gls{lidar} data from an overhead flight, we detect trees with a Local Maximum Filter on the $z$ axis, i.e., we assume that the trees' crowns are the highest observed objects in a forest, and all local maxima are considered tree detections. These detections are used to position the tree models in our simulated environment. This environment covers around 0.77 $\text{km}^2$ (190 acres) and contains 10,692 trees. Using the overhead data as a reference, the user draws polygons to specify the region of interest, and the coverage algorithm will then compute a coverage path as an ordered set of waypoints $\mathcal{G}$. This is illustrated in \cref{fig:coverage-planning}. We run SLOAM with no pose noise to generate a semantic map as the robot flies for $\sim$10km autonomously. The final tree cylinder count is 10,897.}

% environment density
\gui{The second set of experiments measures the impact of environment density on the autonomous flight stack, providing insight to the system's potential to perform fast flights as well as the design considerations of autonomous \glspl{uav}. We use tree positions from a real forest and generate ten simulated forests with the same distribution but different densities (by scaling up or down the x and y positions). Taking the \gls{uav} and tree trunks' diameters into account, the average gap between trees for the densest forest is 2.19 meters.} These simulated experiments are performed with the maximum velocity set as 11.0 m/s and without pose estimation noise. Four missions with a combined trajectory length of $\sim$1 km are sent to the robot for each environment.

From \cref{fig:density-vs-velocity1}, there is a clear trend showing that the average velocity decreases with increasing tree density. Fig.~\ref{fig:density-vs-length} shows that the actual distance traveled by the \gls{uav} increases with tree density due to obstacle avoidance. This, combined with the decrease of average velocity, leads to a more drastic increase in time to complete mission, as shown in the red curve. These experiments not only show our autonomous flight system's fast flight capabilities, but also indicate that the difficulty of accomplishing flights with a limited battery life increases rapidly with the density of the environment. Moreover, under a dense canopy, the \gls{uav}'s average velocity is limited by the environment instead of its maximum design velocity. Thus, flight time is the dominant factor for the mission range. This supports our choice of using a Li-Ion battery over a Li-Po battery because the former has higher specific energy (Wh/kg).

\subsection{Real-world experiments and analysis}
\label{sec:result_and_analysis_real_world}

\begin{figure}[!ht]
    \vspace{-0.1in}
        \centering
            \includegraphics[trim=4cm 1cm 4cm 2.5cm, clip, width=0.85\columnwidth]{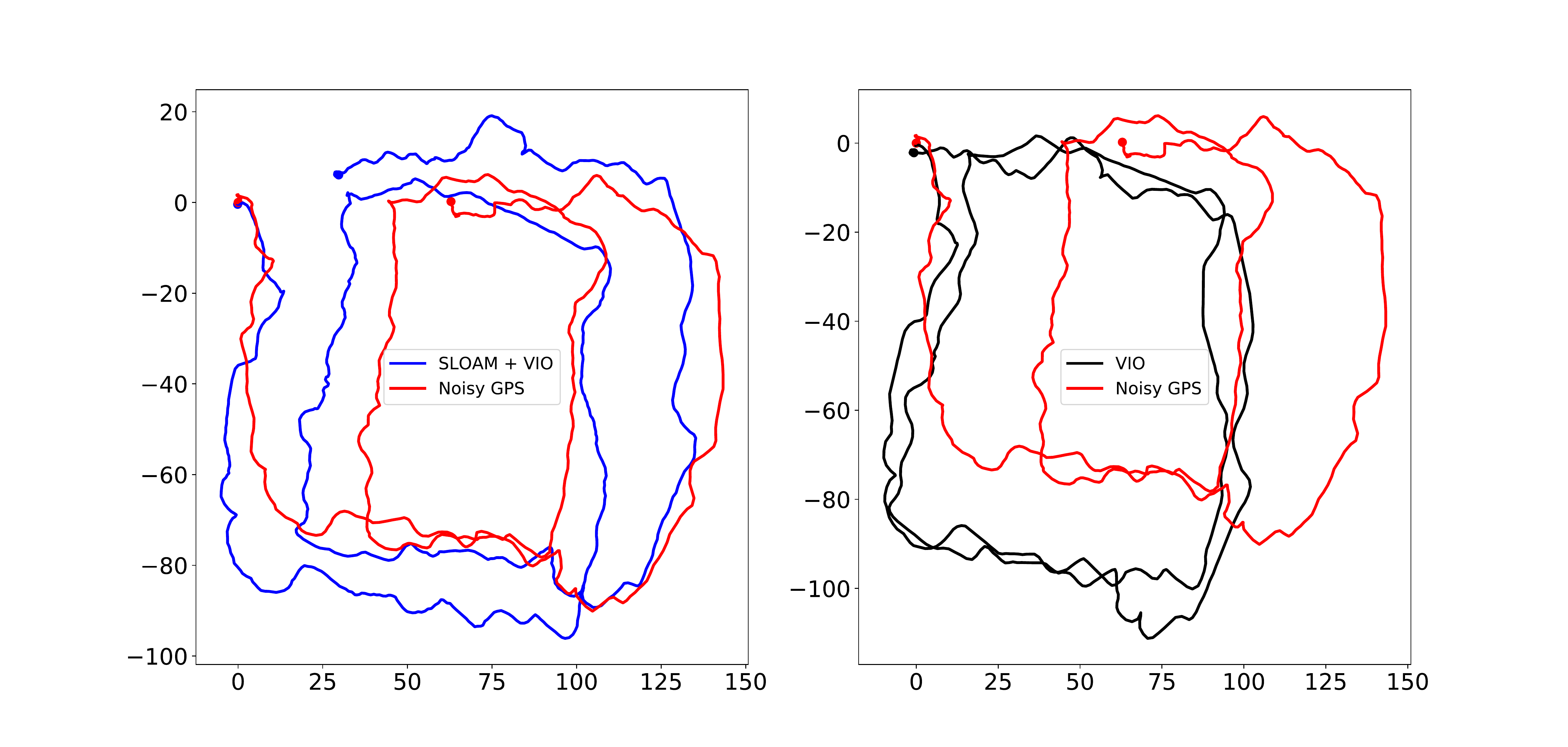}
            \caption{\textbf{Top-down view of the 800 m trajectory.} In this experiment, the autonomy stack relies only on \gls{vio} to control the \gls{uav}. The objective is to perform two square loops and return home. While according to VIO the robot is successful, both SLOAM and noisy GPS show that the \gls{uav} drifts significantly. The GPS sensor reported a $\sim$10 \text{m} standard deviation for X and Y position estimates. \xu{However, as the GPS does not drift over time, it provides a useful high-level reference of the global position. Therefore, here we use it for qualitative evaluation.}}
    \label{fig:top-down-gps}
        \vspace{-0.25in}
\end{figure}

\begin{figure}[!ht]
    % \vspace{-0.05in}
        \centering
            \includegraphics[trim=4cm 1cm 4cm 2cm, clip, width=0.75\columnwidth]{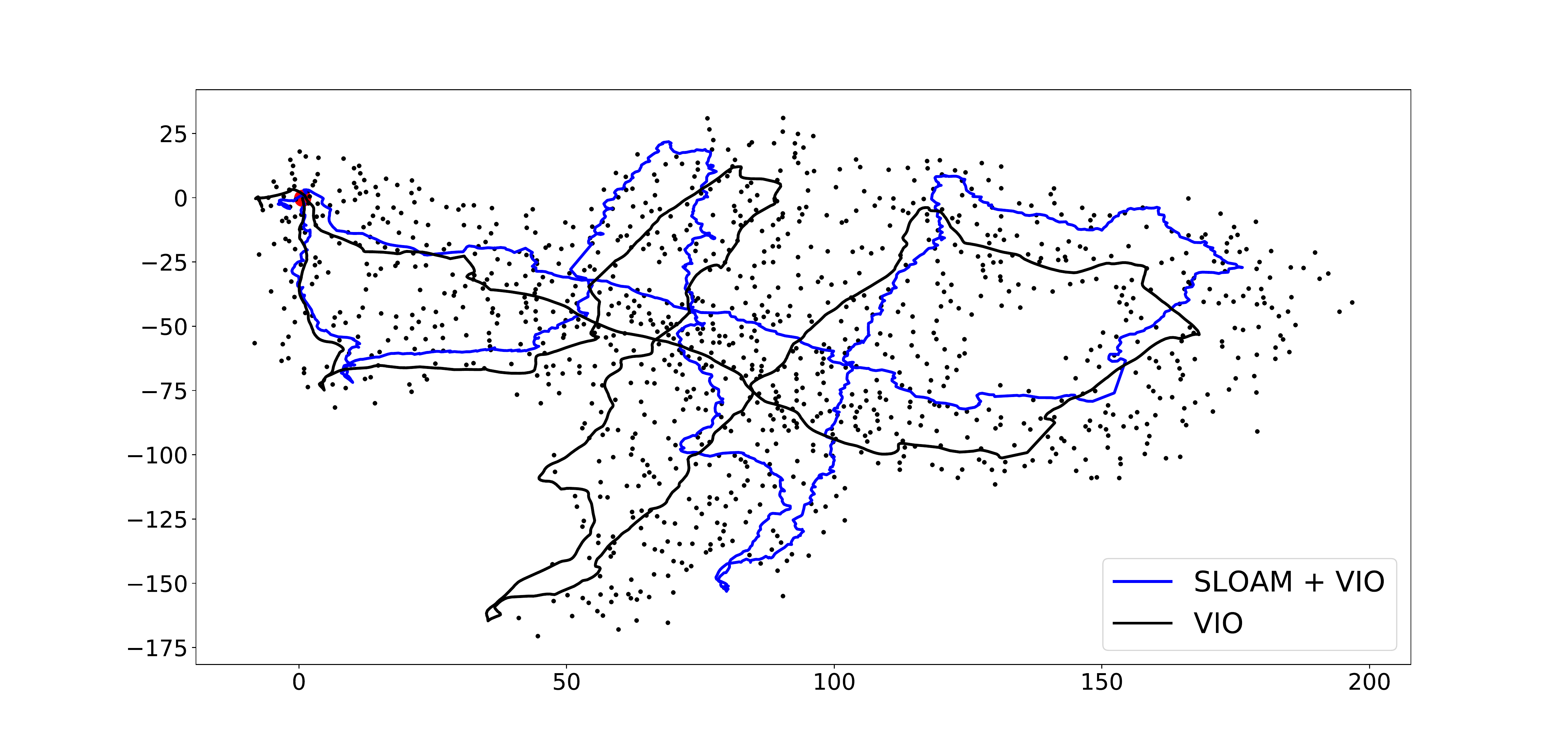}
            \caption{\textbf{Top-down view of the 1.1 km trajectory.} In this experiment, the UAV is manually piloted to guarantee returning to take-off position, so that the drift can be quantified. The black dots represent the tree trunks detected by our mapping algorithm.}
    \label{fig:top-down-loop}
        \vspace{-0.10in}
\end{figure}

In the first field experiment, our goal is to show that relying only on \gls{vio} will lead to failure in completing the mission. As illustrated in~\xu{\cref{fig:top-down-gps}}, the autonomy stack relies only on \gls{vio} to control the \gls{uav}. The commanded mission is to perform two square loops of $\sim$800 m and return home. According to \gls{vio}, the robot is successful, but according to SLOAM and GPS, the \gls{vio} drifts significantly ($\sim$60 m). SLOAM and GPS agree with each other within a 10$\sim$20 m margin. A possible reason why \gls{vio} drifts significantly in this experiment is that the \gls{uav}'s motion is smooth and close to a constant velocity state, which may have formed degenerate cases for \gls{vio}~\cite{yang2019degenerate-vio-degenerate}.

% autonomous 
In the second field experiment, we demonstrate that the \gls{vio} drift can be corrected with semantic SLAM. To quantify the drift, we manually pilot the \gls{uav} so that the takeoff and landing positions are exactly the same. As shown in~\cref{fig:top-down-loop}, the trajectory is a loop of $\sim$1.1 km, covering most of a 200 $m^2$ area while detecting and modeling $1157$ trees. The GPS was not able to get a lock during this experiment. Therefore, we cannot include it in the results of this test.

In Table~\ref{tab:drift}, we compare the accumulated drift based on the starting and ending positions of the pose estimated by \gls{vio} and SLOAM + \gls{vio}. Although stereo \gls{vio} outputs consistent local estimates, it is unreliable over long-range flights, especially along the Z direction. Directly using such estimates will lead to the robot deviating from the desired trajectory. SLOAM benefits from the local consistency of \gls{vio} but can reduce the drift by using landmarks and the ground model. Most of the Z-axis drift is corrected by using our semantic \gls{lidar} framework, and the XY drift is reduced. \gui{Other \gls{lidar} state estimators such as LIO-SAM~\cite{liosam2020shan} and Fast-Lio2~\cite{xu2021fast} were also tested, but due to a~\gls{lidar} connector malfunction that caused intermittent data loss, these methods failed. This is also a good example of why having sensor redundancy is key to long-range operation, since a hardware malfunction such as this can happen anytime to a robot system.}

\begin{table}[h]
\centering
\caption{Distance of the final estimated pose from the beginning of the trajectory on the 1.1 km loop.}
\label{tab:drift}
\begin{tabular}{lccc}
Method      & XY Drift (m) & Z Drift (m) & Total Drift (m) \\\hline

VIO         & 7.92         & -6.27       & 10.10           \\
SLOAM + VIO & 3.93         & 0.71        & 3.99            
\end{tabular}
\vspace{-0.225in}
\end{table}

\subsection{Discussion}
Navigating in forests is difficult, especially when the density is high. Even with the same planner configuration, the average speed, mission time and length vary significantly. There is a tight connection between perception and the decision-making loop. If there is drift in state estimation, the planning will deviate and fail to execute the mission; additionally, the state estimation results can be influenced by the \gls{uav}'s motion.

Traditional odometry systems are good at outputting smooth estimation over a moderate distance,  but semantic SLAM can reduce drift for longer flights or different environment densities. Therefore, using semantics to compensate for the drift in \gls{vio} results in greater robustness and reliability. Moreover, different sensors have different degenerate cases and failure modes. Forest-like environments aggravate this due to perception aliasing, excessive heat, humidity, sunlight, and dust. Therefore, sensor redundancy can increase reliability.

A \gls{sam} framework such as the one proposed by LIO-SAM~\cite{liosam2020shan} can be extended to leverage our semantic map and could further reduce the global drift from SLOAM. However, the computation demand for handling loop closures with large maps is high. %, which 
%will impact the performance of the autonomy stack given the limited computational power on the \gls{uav}. 
\guitocheck{Thus, the use of \gls{sam} at for large-scale semantic maps is left as future work.}
    
    \section{Conclusion and Future Work}
\label{sec:conclusion}

Generating semantic maps of forests and orchards is important for understanding the magnitude of the carbon challenge and to develop new strategies for precision agriculture. Semantic mapping requires under-canopy flight and measurements at the individual tree or plant level. In this paper, we described the key hardware and software components required for an autonomous \gls{uav} to build large-scale maps of unstructured forests in a GPS-challenged environment with varying illumination (sunlight, shadows) and wind, without any human intervention. These include the use of cameras, an IMU, and a \gls{lidar} for state estimation; the segmentation of trees and ground planes using a \gls{lidar}; and the real-time integration of the semantic map in the control and planning loop to minimize drift and to facilitate localization, mapping, and planning. To our knowledge, this is the first time that real-time semantic SLAM has been integrated into the feedback loop for autonomous \gls{uav} navigation, and detailed, multi-resolution maps over a large area have been built in an automated fashion, using only onboard sensing and computation. We hope our work can inspire future research in not only SLAM but also aerial autonomy. Future work involves incorporating the semantic map to actively choose actions that reduces uncertainty of the information gathered by the system.

    \section{Acknowledgment}
This work was supported by funding from the IoT4Ag Engineering Research Center~\cite{iot4ag} funded by the National Science Foundation (NSF) under NSF Cooperative Agreement Number EEC-1941529, and C-BRIC, a Semiconductor Research Corporation Joint University Microelectronics Program program cosponsored by DARPA. This work was also partially supported by INCT-INSac grants CNPq 465755/2014-3, FAPESP 2014/50851-0, and 2017/17444-0. We gratefully acknowledge the Distributed and Collaborative Intelligent Systems and Technology Collaborative Research Alliance (DCIST) team for building the simulation infrastructure, and New Jersey State Forestry Services staff for supporting experiments in the Wharton State Forest. We also thank Dr. Avraham Cohen for the original design of Falcon 4 platform, Dr. Ke Sun and Dr. Kartik Mohta for their insights and help regarding autonomous flight system development and field experiments, and Dr. Luiz Rodriguez for the over-canopy LiDAR data.

\bibliographystyle{IEEEtran}
\bibliography{ref, ag-survey-ref}

\begin{thebibliography}{10}
\providecommand{\url}[1]{#1}
\csname url@rmstyle\endcsname
\providecommand{\newblock}{\relax}
\providecommand{\bibinfo}[2]{#2}
\providecommand\BIBentrySTDinterwordspacing{\spaceskip=0pt\relax}
\providecommand\BIBentryALTinterwordstretchfactor{4}
\providecommand\BIBentryALTinterwordspacing{\spaceskip=\fontdimen2\font plus
\BIBentryALTinterwordstretchfactor\fontdimen3\font minus
  \fontdimen4\font\relax}
\providecommand\BIBforeignlanguage[2]{{%
\expandafter\ifx\csname l@#1\endcsname\relax
\typeout{** WARNING: IEEEtran.bst: No hyphenation pattern has been}%
\typeout{** loaded for the language `#1'. Using the pattern for}%
\typeout{** the default language instead.}%
\else
\language=\csname l@#1\endcsname
\fi
#2}}

\bibitem{mohta2018experiments}
K.~Mohta, K.~Sun, S.~Liu, M.~Watterson, B.~Pfrommer, J.~Svacha, Y.~Mulgaonkar,
  C.~J. Taylor, and V.~Kumar, ``Experiments in fast, autonomous, gps-denied
  quadrotor flight,'' in \emph{2018 IEEE International Conference on Robotics
  and Automation (ICRA)}.\hskip 1em plus 0.5em minus 0.4em\relax IEEE, 2018,
  pp. 7832--7839.

\bibitem{quigley2019open}
M.~Quigley, K.~Mohta, S.~S. Shivakumar, M.~Watterson, Y.~Mulgaonkar,
  M.~Arguedas, K.~Sun, S.~Liu, B.~Pfrommer, and V.~Kumar, ``The open vision
  computer: An integrated sensing and compute system for mobile robots,'' in
  \emph{2019 International Conference on Robotics and Automation (ICRA)}.\hskip
  1em plus 0.5em minus 0.4em\relax IEEE, 2019, pp. 1834--1840.

\bibitem{forest-gps3-carreiras2013estimating}
J.~Carreiras, J.~B. Melo, and M.~J. Vasconcelos, ``Estimating the above-ground
  biomass in miombo savanna woodlands (mozambique, east africa) using l-band
  synthetic aperture radar data,'' \emph{Remote Sensing}, vol.~5, no.~4, pp.
  1524--1548, 2013.

\bibitem{bowman2017probabilistic}
S.~L. Bowman, N.~Atanasov, K.~Daniilidis, and G.~J. Pappas, ``Probabilistic
  data association for semantic slam,'' in \emph{2017 IEEE international
  conference on robotics and automation (ICRA)}.\hskip 1em plus 0.5em minus
  0.4em\relax IEEE, 2017, pp. 1722--1729.

\bibitem{kostavelis2015semantic}
I.~Kostavelis and A.~Gasteratos, ``Semantic mapping for mobile robotics tasks:
  A survey,'' \emph{Robotics and Autonomous Systems}, vol.~66, pp. 86--103,
  2015.

\bibitem{nicholson2018quadricslam}
L.~Nicholson, M.~Milford, and N.~S{\"u}nderhauf, ``Quadricslam: Dual quadrics
  from object detections as landmarks in object-oriented slam,'' \emph{IEEE
  Robotics and Automation Letters}, vol.~4, no.~1, pp. 1--8, 2018.

\bibitem{yang2019cubeslam}
S.~Yang and S.~Scherer, ``Cubeslam: Monocular 3-d object slam,'' \emph{IEEE
  Transactions on Robotics}, vol.~35, no.~4, pp. 925--938, 2019.

\bibitem{bavle2020vps}
H.~Bavle, P.~De~La~Puente, J.~P. How, and P.~Campoy, ``Vps-slam: visual planar
  semantic slam for aerial robotic systems,'' \emph{IEEE Access}, vol.~8, pp.
  60\,704--60\,718, 2020.

\bibitem{murali2017utilizing}
V.~Murali, H.-P. Chiu, S.~Samarasekera, and R.~T. Kumar, ``Utilizing semantic
  visual landmarks for precise vehicle navigation,'' in \emph{2017 IEEE 20th
  International Conference on Intelligent Transportation Systems (ITSC)}.\hskip
  1em plus 0.5em minus 0.4em\relax IEEE, 2017, pp. 1--8.

\bibitem{ok2019robust}
K.~Ok, K.~Liu, K.~Frey, J.~P. How, and N.~Roy, ``Robust object-based slam for
  high-speed autonomous navigation,'' in \emph{2019 International Conference on
  Robotics and Automation (ICRA)}.\hskip 1em plus 0.5em minus 0.4em\relax IEEE,
  2019, pp. 669--675.

\bibitem{bavle2018stereo}
H.~Bavle, S.~Manthe, P.~De~La~Puente, A.~Rodriguez-Ramos, C.~Sampedro, and
  P.~Campoy, ``Stereo visual odometry and semantics based localization of
  aerial robots in indoor environments,'' in \emph{2018 IEEE/RSJ International
  Conference on Intelligent Robots and Systems (IROS)}.\hskip 1em plus 0.5em
  minus 0.4em\relax IEEE, 2018, pp. 1018--1023.

\bibitem{lin2018autonomous}
Y.~Lin, F.~Gao, T.~Qin, W.~Gao, T.~Liu, W.~Wu, Z.~Yang, and S.~Shen,
  ``Autonomous aerial navigation using monocular visual-inertial fusion,''
  \emph{Journal of Field Robotics}, vol.~35, no.~1, pp. 23--51, 2018.

\bibitem{oleynikova2020open}
H.~Oleynikova, C.~Lanegger, Z.~Taylor, M.~Pantic, A.~Millane, R.~Siegwart, and
  J.~Nieto, ``An open-source system for vision-based micro-aerial vehicle
  mapping, planning, and flight in cluttered environments,'' \emph{Journal of
  Field Robotics}, vol.~37, no.~4, pp. 642--666, 2020.

\bibitem{zhou2020ego}
X.~Zhou, Z.~Wang, H.~Ye, C.~Xu, and F.~Gao, ``Ego-planner: An esdf-free
  gradient-based local planner for quadrotors,'' \emph{IEEE Robotics and
  Automation Letters}, vol.~6, no.~2, pp. 478--485, 2020.

\bibitem{mohta2018fast}
K.~Mohta, M.~Watterson, Y.~Mulgaonkar, S.~Liu, C.~Qu, A.~Makineni, K.~Saulnier,
  K.~Sun, A.~Zhu, J.~Delmerico, and V.~Kumar, ``Fast, autonomous flight in
  gps-denied and cluttered environments,'' \emph{Journal of Field Robotics},
  vol.~35, no.~1, pp. 101--120, 2018.

\bibitem{chen2020sloam}
S.~W. Chen, G.~V. Nardari, E.~S. Lee, C.~Qu, X.~Liu, R.~A.~F. Romero, and
  V.~Kumar, ``Sloam: Semantic lidar odometry and mapping for forest
  inventory,'' \emph{IEEE Robotics and Automation Letters}, vol.~5, no.~2, pp.
  612--619, 2020.

\bibitem{milioto2019iros-rangenet++}
A.~Milioto, I.~Vizzo, J.~Behley, and C.~Stachniss, ``{RangeNet++: Fast and
  Accurate LiDAR Semantic Segmentation},'' in \emph{IEEE/RSJ Intl.~Conf.~on
  Intelligent Robots and Systems (IROS)}, 2019.

\bibitem{sun2018robust}
K.~Sun, K.~Mohta, B.~Pfrommer, M.~Watterson, S.~Liu, Y.~Mulgaonkar, C.~J.
  Taylor, and V.~Kumar, ``Robust stereo visual inertial odometry for fast
  autonomous flight,'' \emph{IEEE Robotics and Automation Letters}, vol.~3,
  no.~2, pp. 965--972, 2018.

\bibitem{bahnemann2021revisiting}
R.~B{\"a}hnemann, N.~Lawrance, J.~J. Chung, M.~Pantic, R.~Siegwart, and
  J.~Nieto, ``Revisiting boustrophedon coverage path planning as a generalized
  traveling salesman problem,'' in \emph{Field and Service Robotics}.\hskip 1em
  plus 0.5em minus 0.4em\relax Springer, 2021, pp. 277--290.

\bibitem{shan2018lego}
T.~Shan and B.~Englot, ``Lego-loam: Lightweight and ground-optimized lidar
  odometry and mapping on variable terrain,'' in \emph{2018 IEEE/RSJ
  International Conference on Intelligent Robots and Systems (IROS)}.\hskip 1em
  plus 0.5em minus 0.4em\relax IEEE, 2018, pp. 4758--4765.

\bibitem{mellinger2011minimum}
D.~Mellinger and V.~Kumar, ``Minimum snap trajectory generation and control for
  quadrotors,'' in \emph{2011 IEEE international conference on robotics and
  automation}.\hskip 1em plus 0.5em minus 0.4em\relax IEEE, 2011, pp.
  2520--2525.

\bibitem{liu2018search}
S.~Liu, K.~Mohta, N.~Atanasov, and V.~Kumar, ``Search-based motion planning for
  aggressive flight in {SE}(3),'' \emph{IEEE Robotics and Automation Letters},
  vol.~3, no.~3, pp. 2439--2446, 2018.

\bibitem{lee2010geometric}
T.~Lee, M.~Leok, and N.~H. McClamroch, ``Geometric tracking control of a
  quadrotor uav on se (3),'' in \emph{49th IEEE conference on decision and
  control (CDC)}.\hskip 1em plus 0.5em minus 0.4em\relax IEEE, 2010, pp.
  5420--5425.

\bibitem{yang2019degenerate-vio-degenerate}
Y.~Yang, P.~Geneva, K.~Eckenhoff, and G.~Huang, ``Degenerate motion analysis
  for aided ins with online spatial and temporal sensor calibration,''
  \emph{IEEE Robotics and Automation Letters}, vol.~4, no.~2, pp. 2070--2077,
  2019.

\bibitem{liosam2020shan}
T.~Shan, B.~Englot, D.~Meyers, W.~Wang, C.~Ratti, and R.~Daniela, ``Lio-sam:
  Tightly-coupled lidar inertial odometry via smoothing and mapping,'' in
  \emph{IEEE/RSJ International Conference on Intelligent Robots and Systems
  (IROS)}.\hskip 1em plus 0.5em minus 0.4em\relax IEEE, 2020, pp. 5135--5142.

\bibitem{xu2021fast}
W.~Xu and F.~Zhang, ``Fast-lio: A fast, robust lidar-inertial odometry package
  by tightly-coupled iterated kalman filter,'' \emph{IEEE Robotics and
  Automation Letters}, vol.~6, no.~2, pp. 3317--3324, 2021.

\bibitem{iot4ag}
\BIBentryALTinterwordspacing
{The Internet of Things for Precision Agriculture (IoT4Ag), an NSF Engineering
  Research Center}. [Online]. Available: \url{https://iot4ag.us}
\BIBentrySTDinterwordspacing

\end{thebibliography}

\end{document}